\pdfoutput=1

\documentclass[11pt]{article}

\usepackage[preprint]{acl}

\usepackage{times}
\usepackage{latexsym}
\usepackage{amsfonts}
\usepackage{amsmath}
\usepackage{booktabs}
\usepackage{tikz}
\usepackage{multirow}
\usetikzlibrary{trees}
\usepackage{listings}
\usepackage[T1]{fontenc}

\usepackage[utf8]{inputenc}

\usepackage{microtype}

\usepackage{inconsolata}

\usepackage{graphicx}

\title{Enhancing Retrieval Systems with Inference-Time Logical Reasoning}

\author{\textbf{Felix Faltings\textsuperscript{1}\thanks{Work done during an internship at Accenture.}},
\textbf{Wei Wei\textsuperscript{2}},
\textbf{Yujia Bao\textsuperscript{2}}
\\
 \textsuperscript{1}MIT,
 \textsuperscript{2}Center for Advanced AI, Accenture
\\
\texttt{faltings@mit.edu}, \texttt{\{wei.h.wei, yujia.bao\}@accenture.com}
   }

\newcommand{\AND}{\ifmmode \text{ AND } \else AND \fi}
\newcommand{\OR}{\ifmmode \text{ OR } \else OR \fi}
\newcommand{\NOT}{\ifmmode \text{ NOT } \else NOT \fi}
\newcommand{\OPAND}{\text{OP}_{AND}}
\newcommand{\OPOR}{\text{OP}_{OR}}
\newcommand{\OPNOT}{\text{OP}_{NOT}}
\newcommand{\Method}{ITLR}

\begin{document}
\maketitle
\begin{abstract}
Traditional retrieval methods rely on transforming user queries into vector representations and retrieving documents based on cosine similarity within an embedding space. While efficient and scalable, this approach often fails to handle complex queries involving logical constructs such as negations, conjunctions, and disjunctions. In this paper, we propose a novel inference-time logical reasoning framework that explicitly incorporates logical reasoning into the retrieval process. Our method extracts logical reasoning structures from natural language queries and then composes the individual cosine similarity  scores to formulate the final document scores. This approach enables the retrieval process to handle complex logical reasoning without compromising computational efficiency. Our results on both synthetic and real-world benchmarks demonstrate that the proposed method consistently outperforms traditional retrieval methods across different models and datasets, significantly improving retrieval performance for complex queries.

\end{abstract}

\section{Introduction}

Retrieval systems are integral to many applications, including search engines, question-answering systems, and recommendation platforms \citep{baeza1999modern,lewis2020retrieval,gao2023retrieval}. Modern systems operate by transforming user queries into vector representations and retrieving documents based on cosine similarity within an embedding space \cite{reimers2019sentence, wang2023improving,zhao2024dense, lee2024nv}. This approach is highly efficient and scalable, as cosine similarity computations are fast and can handle large-scale data. However, the reliance on cosine similarity and static embeddings often limits the system's ability to understand and process complex queries that involve logical constructs such as negations.

\begin{figure}[t!]
    \centering
    \includegraphics[height=1.8in]{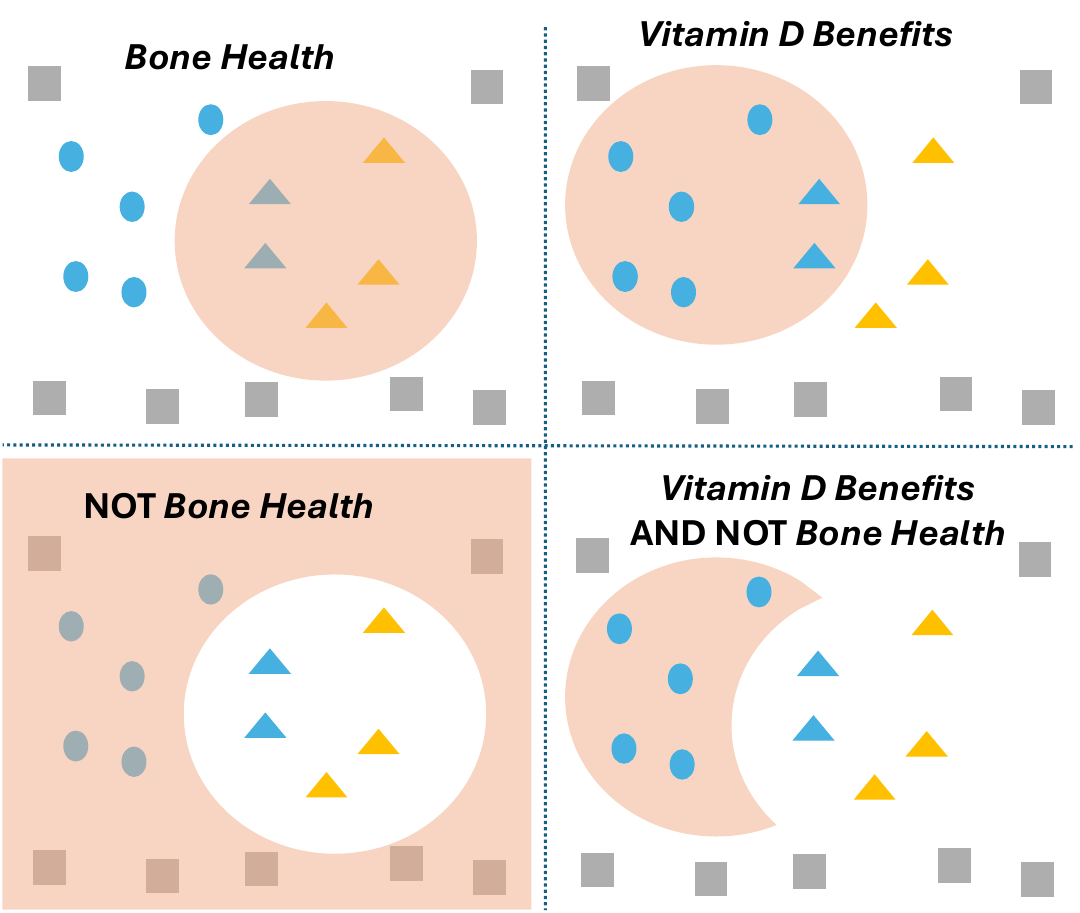}
    \caption{Given a query ``What are the benefits of vitamin D, focusing on benefits other than bone health?", we first convert the query into the logical expression ``Vitamin D Benefits AND NOT Bone Health". We then calculate the cosine similarity scores for each term (top row) and combine these scores to generate the final results.}
    \label{fig:diagram}
    \vspace{-15pt}
\end{figure}

Large Language Models (LLMs) have demonstrated remarkable capabilities in inference-time reasoning \citep{wei2022chain,yao2024tree}. Recently,
\cite{jiang2023structgpt,gu2022don,sun2023think,luo2023chatkbqa} have successfully applied LLM's reasoning capability to improve the retrieval performance for knowledge-based question answering, however, the application of inference-time reasoning for general retrieval systems remains relatively unexplored.

We posit that integrating logical reasoning at inference time is equally crucial for retrieval tasks, especially when dealing with complex queries that cannot be accurately represented using simple similarity measures~\citep{meghini1993model, ounis1998relief}. Consider a query like "What are the benefits of vitamin D, focusing on benefits other than bone health?" A traditional retrieval system, relying solely on cosine similarity, may struggle to exclude documents related to bone health due to the embedding's inability to represent the negation effectively. The system tends to retrieve documents that are globally similar to the query, failing to account for specific logical instructions, such as exclusions or combinations of concepts.

To address this limitation, we propose a novel inference-time reasoning framework for retrieval systems that explicitly incorporates logical reasoning into the retrieval process. Our approach involves three key steps. First, \textbf{Logical Query Transformation}: We utilize an LLM to parse and rewrite the natural language query into a logical form, such as "A AND B OR NOT C," where A, B, and C represent different semantic components of the query. Second, \textbf{Term Embedding and Similarity Computation}: Each term identified in the logical form (A, B, C, etc.) is individually encoded into the embedding space. We compute the cosine similarity between each term's embedding and the embeddings of documents in the corpus. Third, \textbf{Score Composition Based on Logical Relations}: We combine the similarity scores of each term in accordance with their logical relations.
Our approach adds minimal overhead since the embedding can be performed in parallel.


We validate our approach through comprehensive experiments. We first create synthetic datasets with queries of varying logical complexity to test the limitations of existing retrieval algorithms. Our findings indicate that as the number of logical terms increases, the performance of traditional retrieval methods degrades significantly, while our method better maintains performance, demonstrating robustness against query complexity. We also evaluated our algorithm on three real-world datasets: NFCorpus~\citep{boteva2016}, SciFact~\citep{wadden-etal-2020-fact} and ArguAna~\citep{thakur2021beir}. Specifically, we augmented these three datasets with natural language queries that target compositional reasoning. We tested our method using four commonly used embedding models. The results show that our approach consistently outperforms baseline methods across all models and datasets, confirming its effectiveness in practical scenarios.


\section{Method}
Given a natural language query,
\emph{``What are the beneifts of vitamin D, focusing on benefits other than bone health?''},
we first transform it into a logical expression using a pre-trained large language model~\citep{dubey2024llama}, \emph{``Vitamin D Benefits" AND NOT ``Bone Health"}, where the terms in quotes can be any string of text. Given a document, these queries can be seen as logical expressions, which we evaluate in a fuzzy way \cite{novak2012mathematical}, using the scores from a dense retrieval model to assign truth values to each clause. The fuzzy evaluation of the expression then gives a composite retrieval score for the given document. In the following sections, we present the concrete details of our method, starting with the syntax of the logical queries, followed by the retrieval semantics.

\subsection{Query Syntax}

Queries are made up of terms--which can be any string of text--combined using operators. We allow three operators, AND, OR, and NOT. Formally, the syntax of the language is described by the following simple grammar,
\begin{align*}
    &T \to U \OR U \,|\, U\\
    &U \to V \AND V \,|\, V\\
    &V \to \NOT W \,|\, W \\
    &W \to \textit{string} \,|\, (T)
\end{align*}

where the use of distinguished non-terminals, $T,U,V$, and $W$ enforces an operator priority, $\NOT \succ \AND \succ \OR$, which is itself overridden by parentheses.

\subsection{Query Semantics}
For each term $t_j$ in a query, and each document $D_i$ in a corpus, we compute a score $s_{ji}$ using the dense retrieval model. Usually, this is the cosine similarity between the embedding vectors of the term and document. The semantics of the query then tell us how to combine the scores $s_{ji}$ into a single score $s_i$ which we can use for retrieval.

Consider a query of the form,
\begin{equation*}
    (t_1 \OR t_2) \AND \NOT t_3.
    \label{eqn:example}
\end{equation*}
Then, for document $D_i$, if $s_{1i},s_{2i},$ and $s_{3i}$ are the scores obtained from the dense retrieval model, we take the composed retrieval score to be,
\begin{equation*}
    s_i = \OPAND(\OPOR(s_{1i}, s_{2i}), \OPNOT(s_{3i})),
\end{equation*}
where $\OPAND$, $\OPOR$ and $\OPNOT$ are functions that define how scores should be combined depending on the query operator. We detail our choice of operators in the next section.

\begin{figure}[!ht]
    \centering
    \begin{minipage}{0.225\textwidth}
        \centering
        \begin{tikzpicture}[
  edge from parent/.style={draw, thick, edge from parent fork down},
  level 1/.style={sibling distance=2cm, level distance=1cm},
  level 2/.style={sibling distance=1.5cm, level distance=1cm}
  ]
  \node {AND}
    child { node {OR}
      child { node {$t_1$}}
      child { node {$t_2$}}
    }
    child { node {NOT}
      child { node {$t_3$}}
    };
\end{tikzpicture}
    \end{minipage}%
    \begin{minipage}{0.225\textwidth}
        \centering
        \begin{tikzpicture}[
  edge from parent/.style={draw, thick, edge from parent fork down},
  level 1/.style={sibling distance=2cm, level distance=1cm},
  level 2/.style={sibling distance=1.5cm, level distance=1cm}
  ]
  \node {$\OPAND$}
    child { node {$\OPOR$}
      child { node {$s_{1i}$}}
      child { node {$s_{2i}$}}
    }
    child { node {$\OPNOT$}
      child { node {$s_{3i}$}}
    };
\end{tikzpicture}
    \end{minipage}
    \vspace{-10pt}
    \caption{Example parse tree (left) and corresponding graph of operations (right).}
    \label{fig:parse_tree}
    \vspace{-10pt}
\end{figure}
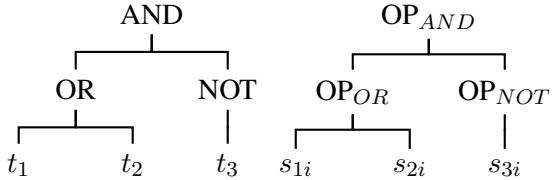

In general, each query can be parsed using the grammar described above, resulting in a parse tree, which is directly translated into a tree of operations acting on the scores $s_{1i},s_{2i}$, and $s_{3i}$ as shown in Fig.~\ref{fig:parse_tree}. A more complicated example is given in the appendix. This tells us how to compute the final retrieval score $s_i$. More formally, this could be written as an attribute grammar \citep{knuth2005genesis}.

\subsection{Score Operations}
\label{sec:score_operations}
The choice of the operators $\OPAND$, $\OPOR$, $\OPNOT$ should reflect the logical semantics of the query. For example, for the conjunction $t_1 \AND t_2$, with scores $s_{1i}$ and $s_{2i}$, the composite score should be low if \textit{any} of the two scores is low. We consider the following choices of operator,
\begin{align*}
    &\OPAND(x,y) = x*y\,|\,x+y\,|\,\min(x,y)\\
    &\OPOR(x,y) = x+y\,|\,\max(x,y)\\
    &\OPNOT(x,y) = 1-x\,|\,1/x
\end{align*}
We evaluate all combinations of these operators in our experiments. Our default choice is
$\OPAND(x,y)=x*y$, $\OPOR(x,y)=x+y$, and $\OPNOT(x,y)=(1-x)$, which we find to work best.

\section{Results}
We start by validating the performance of the logical retrieval system itself on synthetic data. Next, we assess the system's utility for retrieval on real data. In all our experiments, we evaluate using three base embedding models: Nvidia's \texttt{NV-Embed-V1} \cite{lee2024nv}, Mistral's \texttt{nv-embedqa-mistral-7b-v2} and OpenAI's \texttt{text-embedding-v3-large} and \texttt{text-embedding-v3-small}.

\subsection{Synthetic Data}

\paragraph{Three Term Queries}
We first evaluate performance on all possible queries formed of three terms using synthetic data. This gives 32 ``templates", such as,
\begin{equation*}
    t_1 \AND t_2 \OR \NOT t_3
\end{equation*}
The three placeholders $t_1,t_2$, and $t_3$ are filled in by terms. For each possible template, we generate 100 queries by filling in the placeholders with random topics from a set generated by Llama3-70b. For each query we then generate documents that match and don't match the query with Llama3-70b. For example, for the query,
\begin{equation*}
    \text{``dog" AND ``cat" AND ``mouse''},
\end{equation*}
we generate one document that matches, which is related to all three terms, and three documents that don't match, which are related to all but one of the terms. See Appendix~\ref{sec:app_synth_data} for more details.

The results are presented in Table~\ref{tab:synth_data}. We report the standard nDCG@10 in all our results. We show the results when passing the query directly to the retrieval model (base) vs. composing the retrieval model scores for each term (logical). For reference, the performance when using random scores is around 0.7. We see that logical retrieval outperforms the baseline, with the most gains coming from queries with negations. We did not see large differences between embedding models. See Table~\ref{tab:synth_data_embd_model} in the appendix for a breakdown.

\begin{table}
\centering
\small
\setlength\tabcolsep{4mm}
\begin{tabular}{lrrrr}
\toprule
& \multicolumn{4}{c}{Number of negations} \\\cmidrule{2-5}
 & 0 & 1 & 2 & 3 \\
 \midrule
Base & 0.95 & 0.77 & 0.65 & 0.52 \\
\Method & \textbf{0.99} & \textbf{0.97} & \textbf{0.96} & \textbf{1.00} \\
\bottomrule
\end{tabular}
\caption{nDCG@10 Results on synthetic data. Dense and logical retrieval systems were evaluated on synthetically generated test cases for all 32 possible logical queries with three terms. We show results broken down by the number of negations in the queries.}
\label{tab:synth_data}
\vspace{-10pt}
\end{table}

\begin{figure}[h!]
    \centering
    \begin{minipage}{0.225\textwidth}
        \centering
        \includegraphics[width=\linewidth]{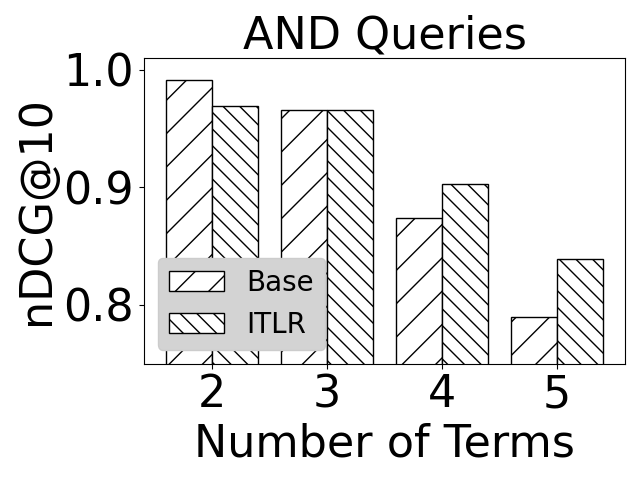}  
    \end{minipage}%
    \begin{minipage}{0.225\textwidth}
        \centering
        \includegraphics[width=\linewidth]{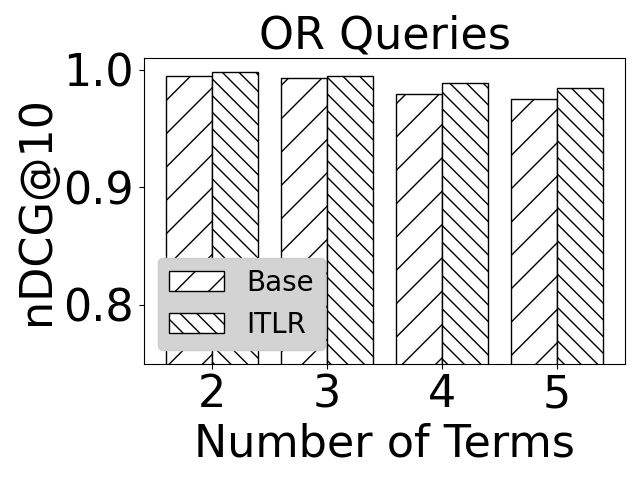}  
    \end{minipage}
    \vspace{-10pt}
    \caption{Performance as the number of terms scales. Baseline dense retrieval and logical retrieval were evaluated on queries connected by \AND and \OR clauses, with increasing number of clauses.}
    \label{fig:and_or}
\end{figure}

\paragraph{Scaling Number of Queries}
We look at performance as the number of terms in the queries scales, focusing this time solely on queries consisting of OR or AND operators. For example,
\begin{equation*}
    \text{``dog" AND ``mouse" AND}\dots\text{ AND ``cat"}
\end{equation*}
We generate data in the same way as before. Our results are presented in Fig.~\ref{fig:and_or}. We see that the gains from logical retrieval increase as the number of terms increases. This is more pronounced for the AND queries than the OR queries, likely since each AND query has a single positive match whereas each OR query has many matches.

\subsection{Operator Combinations}
We tested all combinations of operators $\OPAND$, $\OPOR$ and $\OPNOT$ proposed in section \ref{sec:score_operations} on the three term query data using the \texttt{NV-Embed-V1} embedding model. We present the results in Table~\ref{tab:operator_ablations}. We see that our default choice works best, although the alternative using $\OPAND(x,y)=x+y$ works equally as well. Note that the common choice made in fuzzy logic of $\OPAND(x,y)=\min(x,y)$ and $\OPOR(x,y)=\max(x,y)$ performs quite poorly.

\begin{table}[t]
\small
\centering
\setlength\tabcolsep{4mm}
\begin{tabular}{llrr}
\toprule
 &  & \multicolumn{2}{c}{$\OPNOT$}\\
$\OPAND$ & $\OPOR$ & $1-x$ & $1/x$  \\
\midrule
\multirow[t]{2}{*}{$\min(x,y)$} & $\max(x,y)$ & 0.86 & 0.86 \\
 & $x+y$ & 0.92 & 0.87 \\
\multirow[t]{2}{*}{$x*y$} & $\max(x,y)$ & 0.90 & 0.89 \\
 & $x+y$ & \textbf{0.97} & 0.90 \\
\multirow[t]{2}{*}{$x+y$} & $\max(x,y)$ & 0.91 & 0.81 \\
 & $x+y$ & \textbf{0.97} & 0.82 \\
\bottomrule
\end{tabular}
\caption{nDCG@10 results for logical retrieval on the same data from Table~\ref{tab:synth_data} for different choices of operators, using \texttt{NV-Embed-V1}. Each entry represents a choice for each of the three operators.}
\label{tab:operator_ablations}
\vspace{-10pt}
\end{table}

\subsection{Real Data}
\label{sec:real_data}

Our previous experiments consider very small corpora constructed specifically for each query. We now turn to real datasets. In order to ensure the queries have sufficient compositionality, we generate queries using Llama3-70b. As in our synthetic experiments, we create 3-term logical queries from templates, filled in with topics extracted from the dataset. We create 30 queries per template, resulting in 960 total queries. We ask Llama3-70b to turn these queries into natural language questions and throw away the original queries. We also use Llama3-70b to label the relevance of each document in the corpus to each of the questions. In the appendix we give the prompts that were used and show through examples that the generated queries are realistic.

We compare three methods. The \textbf{Baseline} feeds the question to the dense retrieval model. The \textbf{BRIGHT} baseline first asks Llama3-70b to reason about the question and feeds the reasoning trace to the retrieval model. This is the method used in \cite{su2024bright}. Finally, our \textbf{\Method} method asks Llama3-70b to generate a logical query from the question, which is fed to our logical retrieval system. See the appendix for human evaluation results showing that Llama3-70b is able to successfully formulate logical queries.

We report nDCG@10 results in Table~\ref{tab:real_data}, for three datasets derived respectively from the NFCorpus, SciFact and ArguAna tasks, accessed through MTEB \citep{muennighoff2022mteb}. We see that \Method\ achieves the best performance overall, beating all baselines in a majority of cases. This performance gap becomes larger with more negations, as seen in Table~\ref{tab:real_data_negatives}. See the appendix for a detailed discussion of these cases.

\begin{table}
\setlength\tabcolsep{3.5mm}
\small
\centering
\begin{tabular}{lrrr}
\toprule
& \textbf{NFcorpus} & \textbf{SciFact} & \textbf{ArguAna} \\
\midrule
\multicolumn{3}{l}{\texttt{NV-Embed-V1}:}\\
Baseline & 0.56 & 0.51 & 0.51 \\
BRIGHT & 0.67 & 0.59 & 0.58 \\
\Method & \textbf{0.74} & \textbf{0.64} & 0.64\\
\multicolumn{3}{l}{\texttt{text-embedding-v3-large}:} \\
Base & 0.63 & 0.59 & 0.63 \\
BRIGHT & 0.70 & 0.63 & 0.66 \\
\Method & 0.73 & \textbf{0.64} & \textbf{0.67}\\
\multicolumn{3}{l}{\texttt{text-embedding-v3-small}:}\\
Base & 0.56 & 0.50 & 0.55 \\
BRIGHT & 0.68 & 0.59 & 0.65 \\
\Method & 0.67 & 0.54 & 0.63 \\
\multicolumn{3}{l}{\texttt{nv-embedqa-mistral-7b-v2}:}\\
Base & 0.54 & 0.50 & 0.40 \\
BRIGHT & 0.48 & 0.39 & 0.29 \\
\Method & 0.67 & 0.61 & 0.59 \\
\bottomrule
\end{tabular}
\caption{nDCG@10 Results on real data. For each dataset taken from BEIR \citep{thakur2021beir}, compositional questions were generated using Llama3-70b. We show results for three embedding models and three methods.}
\label{tab:real_data}
\end{table}
\vspace{-2mm}

\begin{table}[t]
\centering
\setlength\tabcolsep{4mm}
\small
\begin{tabular}{lrrrr}
\toprule
& \multicolumn{4}{c}{Number of negations} \\\cmidrule{2-5}
& \textbf{0} & \textbf{1} & \textbf{2} & \textbf{3} \\
\midrule
Base & \textbf{0.81} & 0.60 & 0.51 & 0.36 \\
Reasoning & \textbf{0.81} & 0.68 & 0.64 & 0.56 \\
Logical & 0.76 & \textbf{0.71} &\textbf{ 0.76} & \textbf{0.73} \\
\bottomrule
\end{tabular}
\caption{Breakdown of Table~\ref{tab:real_data} for \texttt{NV-Embed-V1} on NFCorpus by number of negations.}
\label{tab:real_data_negatives}
\vspace{-10pt}
\end{table}

\section{Conclusion}
In this paper, we propose an inference-time logical reasoning framework that addresses the limitations of traditional retrieval methods in managing complex queries with logical constructs. The framework is highly efficient, enabling concurrent computation of retrieval scores for each term. By integrating logical reasoning directly into the retrieval process, our framework consistently outperforms traditional methods on both synthetic and real-world benchmarks, demonstrating particular strength in handling queries with a higher frequency of negations and AND operations.

\section{Limitations}
The logical retrieval system we presented in this paper presents some limitations, as it still underperforms on queries without any negations. We identify some concrete problems that could be addressed in future work. First, in most scenarios, the queries to retrieval systems, such as questions from users, are not given as logical formulas. It is also unreasonable to expect users to write logical formulas on their own. Hence, the system is reliant on reformulating queries into logical queries. While we used a simple prompt to achieve this, it is possible that better performance could be obtained by finetuning a reformulation model. Second, we observed in preliminary experiments that the performance of the system can be improved by calibrating the scores of the underlying retrieval model. For example, when processing AND queries, some terms may receive generally higher retrieval scores than others, biasing retrieval towards documents that match those terms but not the others. We did not find any simple methods to calibrate the scores, but this could be accomplished when training the retrieval model, or by training a calibration model on a large dataset.

\paragraph{Ethical Considerations} We do not foresee any immediate risks of our work as we are not releasing any major new artifacts, such as pre-trained models, which could be used in adverse ways. Retrieval systems are limiting points in applications such as retrieval augmented generation (RAG) since downstream answers are generated based on the documents provided by the retrieval system. Poor retrieval systems may skew the information retrieved from corpora, and thus improving the faithfulness of retrieval systems to queries may be broadly beneficial.

\paragraph{Disclaimer}
This content is provided for general information purposes and is not intended to be used in place of consultation with our professional advisors. This document may refer to marks owned by third parties. All such third-party marks are the property of their respective owners. No sponsorship, endorsement or approval of this content by the owners of such marks is intended, expressed or implied. Copyright © 2024 Accenture. CC BY-NC-ND. All rights reserved. Accenture and its logo are registered trademarks of Accenture.

\bibliography{main}

\begin{thebibliography}{23}
\providecommand{\natexlab}[1]{#1}

\bibitem[{Baeza-Yates et~al.(1999)Baeza-Yates, Ribeiro-Neto et~al.}]{baeza1999modern}
Ricardo Baeza-Yates, Berthier Ribeiro-Neto, et~al. 1999.
\newblock \emph{Modern information retrieval}, volume 463.
\newblock ACM press New York.

\bibitem[{Boteva et~al.(2016)Boteva, Gholipour, Sokolov, and Riezler}]{boteva2016}
Vera Boteva, Demian Gholipour, Artem Sokolov, and Stefan Riezler. 2016.
\newblock \href {http://www.cl.uni-heidelberg.de/~riezler/publications/papers/ECIR2016.pdf} {A full-text learning to rank dataset for medical information retrieval}.

\bibitem[{Dubey et~al.(2024)Dubey, Jauhri, Pandey, Kadian, Al-Dahle, Letman, Mathur, Schelten, Yang, Fan et~al.}]{dubey2024llama}
Abhimanyu Dubey, Abhinav Jauhri, Abhinav Pandey, Abhishek Kadian, Ahmad Al-Dahle, Aiesha Letman, Akhil Mathur, Alan Schelten, Amy Yang, Angela Fan, et~al. 2024.
\newblock The llama 3 herd of models.
\newblock \emph{arXiv preprint arXiv:2407.21783}.

\bibitem[{Gao et~al.(2023)Gao, Xiong, Gao, Jia, Pan, Bi, Dai, Sun, and Wang}]{gao2023retrieval}
Yunfan Gao, Yun Xiong, Xinyu Gao, Kangxiang Jia, Jinliu Pan, Yuxi Bi, Yi~Dai, Jiawei Sun, and Haofen Wang. 2023.
\newblock Retrieval-augmented generation for large language models: A survey.
\newblock \emph{arXiv preprint arXiv:2312.10997}.

\bibitem[{Gu et~al.(2022)Gu, Deng, and Su}]{gu2022don}
Yu~Gu, Xiang Deng, and Yu~Su. 2022.
\newblock Don't generate, discriminate: A proposal for grounding language models to real-world environments.
\newblock \emph{arXiv preprint arXiv:2212.09736}.

\bibitem[{Jiang et~al.(2023)Jiang, Zhou, Dong, Ye, Zhao, and Wen}]{jiang2023structgpt}
Jinhao Jiang, Kun Zhou, Zican Dong, Keming Ye, Wayne~Xin Zhao, and Ji-Rong Wen. 2023.
\newblock Structgpt: A general framework for large language model to reason over structured data.
\newblock \emph{arXiv preprint arXiv:2305.09645}.

\bibitem[{Knuth(2005)}]{knuth2005genesis}
Donald~E Knuth. 2005.
\newblock The genesis of attribute grammars.
\newblock In \emph{Attribute Grammars and their Applications: International Conference WAGA Paris, France, September 19--21, 1990 Proceedings}, pages 1--12. Springer.

\bibitem[{Lee et~al.(2024)Lee, Roy, Xu, Raiman, Shoeybi, Catanzaro, and Ping}]{lee2024nv}
Chankyu Lee, Rajarshi Roy, Mengyao Xu, Jonathan Raiman, Mohammad Shoeybi, Bryan Catanzaro, and Wei Ping. 2024.
\newblock Nv-embed: Improved techniques for training llms as generalist embedding models.
\newblock \emph{arXiv preprint arXiv:2405.17428}.

\bibitem[{Lewis et~al.(2020)Lewis, Perez, Piktus, Petroni, Karpukhin, Goyal, K{\"u}ttler, Lewis, Yih, Rockt{\"a}schel et~al.}]{lewis2020retrieval}
Patrick Lewis, Ethan Perez, Aleksandra Piktus, Fabio Petroni, Vladimir Karpukhin, Naman Goyal, Heinrich K{\"u}ttler, Mike Lewis, Wen-tau Yih, Tim Rockt{\"a}schel, et~al. 2020.
\newblock Retrieval-augmented generation for knowledge-intensive nlp tasks.
\newblock \emph{Advances in Neural Information Processing Systems}, 33:9459--9474.

\bibitem[{Luo et~al.(2023)Luo, Tang, Peng, Guo, Zhang, Ma, Dong, Song, Lin, Zhu et~al.}]{luo2023chatkbqa}
Haoran Luo, Zichen Tang, Shiyao Peng, Yikai Guo, Wentai Zhang, Chenghao Ma, Guanting Dong, Meina Song, Wei Lin, Yifan Zhu, et~al. 2023.
\newblock Chatkbqa: A generate-then-retrieve framework for knowledge base question answering with fine-tuned large language models.
\newblock \emph{arXiv preprint arXiv:2310.08975}.

\bibitem[{Meghini et~al.(1993)Meghini, Sebastiani, Straccia, and Thanos}]{meghini1993model}
Carlo Meghini, Fabrizio Sebastiani, Umberto Straccia, and Costantino Thanos. 1993.
\newblock A model of information retrieval based on a terminological logic.
\newblock In \emph{Proceedings of the 16th annual international ACM SIGIR conference on Research and development in information retrieval}, pages 298--307.

\bibitem[{Muennighoff et~al.(2022)Muennighoff, Tazi, Magne, and Reimers}]{muennighoff2022mteb}
Niklas Muennighoff, Nouamane Tazi, Lo{\"\i}c Magne, and Nils Reimers. 2022.
\newblock Mteb: Massive text embedding benchmark.
\newblock \emph{arXiv preprint arXiv:2210.07316}.

\bibitem[{Nov{\'a}k et~al.(2012)Nov{\'a}k, Perfilieva, and Mockor}]{novak2012mathematical}
Vil{\'e}m Nov{\'a}k, Irina Perfilieva, and Jiri Mockor. 2012.
\newblock \emph{Mathematical principles of fuzzy logic}, volume 517.
\newblock Springer Science \& Business Media.

\bibitem[{Ounis and Pa{\c{s}}ca(1998)}]{ounis1998relief}
Iadh Ounis and Marius Pa{\c{s}}ca. 1998.
\newblock Relief: Combining expressiveness and rapidity into a single system.
\newblock In \emph{Proceedings of the 21st annual international ACM SIGIR conference on research and development in information retrieval}, pages 266--274.

\bibitem[{Reimers(2019)}]{reimers2019sentence}
N~Reimers. 2019.
\newblock Sentence-bert: Sentence embeddings using siamese bert-networks.
\newblock \emph{arXiv preprint arXiv:1908.10084}.

\bibitem[{Su et~al.(2024)Su, Yen, Xia, Shi, Muennighoff, Wang, Liu, Shi, Siegel, Tang et~al.}]{su2024bright}
Hongjin Su, Howard Yen, Mengzhou Xia, Weijia Shi, Niklas Muennighoff, Han-yu Wang, Haisu Liu, Quan Shi, Zachary~S Siegel, Michael Tang, et~al. 2024.
\newblock Bright: A realistic and challenging benchmark for reasoning-intensive retrieval.
\newblock \emph{arXiv preprint arXiv:2407.12883}.

\bibitem[{Sun et~al.(2023)Sun, Xu, Tang, Wang, Lin, Gong, Shum, and Guo}]{sun2023think}
Jiashuo Sun, Chengjin Xu, Lumingyuan Tang, Saizhuo Wang, Chen Lin, Yeyun Gong, Heung-Yeung Shum, and Jian Guo. 2023.
\newblock Think-on-graph: Deep and responsible reasoning of large language model with knowledge graph.
\newblock \emph{arXiv preprint arXiv:2307.07697}.

\bibitem[{Thakur et~al.(2021)Thakur, Reimers, R{\"u}ckl{\'e}, Srivastava, and Gurevych}]{thakur2021beir}
Nandan Thakur, Nils Reimers, Andreas R{\"u}ckl{\'e}, Abhishek Srivastava, and Iryna Gurevych. 2021.
\newblock \href {https://openreview.net/forum?id=wCu6T5xFjeJ} {{BEIR}: A heterogeneous benchmark for zero-shot evaluation of information retrieval models}.
\newblock In \emph{Thirty-fifth Conference on Neural Information Processing Systems Datasets and Benchmarks Track (Round 2)}.

\bibitem[{Wadden et~al.(2020)Wadden, Lin, Lo, Wang, van Zuylen, Cohan, and Hajishirzi}]{wadden-etal-2020-fact}
David Wadden, Shanchuan Lin, Kyle Lo, Lucy~Lu Wang, Madeleine van Zuylen, Arman Cohan, and Hannaneh Hajishirzi. 2020.
\newblock \href {https://doi.org/10.18653/v1/2020.emnlp-main.609} {Fact or fiction: Verifying scientific claims}.
\newblock In \emph{Proceedings of the 2020 Conference on Empirical Methods in Natural Language Processing (EMNLP)}, pages 7534--7550, Online. Association for Computational Linguistics.

\bibitem[{Wang et~al.(2023)Wang, Yang, Huang, Yang, Majumder, and Wei}]{wang2023improving}
Liang Wang, Nan Yang, Xiaolong Huang, Linjun Yang, Rangan Majumder, and Furu Wei. 2023.
\newblock Improving text embeddings with large language models.
\newblock \emph{arXiv preprint arXiv:2401.00368}.

\bibitem[{Wei et~al.(2022)Wei, Wang, Schuurmans, Bosma, Xia, Chi, Le, Zhou et~al.}]{wei2022chain}
Jason Wei, Xuezhi Wang, Dale Schuurmans, Maarten Bosma, Fei Xia, Ed~Chi, Quoc~V Le, Denny Zhou, et~al. 2022.
\newblock Chain-of-thought prompting elicits reasoning in large language models.
\newblock \emph{Advances in neural information processing systems}, 35:24824--24837.

\bibitem[{Yao et~al.(2024)Yao, Yu, Zhao, Shafran, Griffiths, Cao, and Narasimhan}]{yao2024tree}
Shunyu Yao, Dian Yu, Jeffrey Zhao, Izhak Shafran, Tom Griffiths, Yuan Cao, and Karthik Narasimhan. 2024.
\newblock Tree of thoughts: Deliberate problem solving with large language models.
\newblock \emph{Advances in Neural Information Processing Systems}, 36.

\bibitem[{Zhao et~al.(2024)Zhao, Liu, Ren, and Wen}]{zhao2024dense}
Wayne~Xin Zhao, Jing Liu, Ruiyang Ren, and Ji-Rong Wen. 2024.
\newblock Dense text retrieval based on pretrained language models: A survey.
\newblock \emph{ACM Transactions on Information Systems}, 42(4):1--60.

\end{thebibliography}

\appendix

\newpage
\onecolumn

\section{Additional Parsing Example}
Consider the query,
\begin{equation*}
    (\text{``dog"} \OR \text{``cat"} \AND \text{``mouse"}) \AND \NOT \text{``giraffe"} 
\end{equation*}
The corresponding parse tree will be,
\begin{center}
\begin{tikzpicture}[
  edge from parent/.style={draw, thick, edge from parent fork down},
  level 1/.style={sibling distance=2cm, level distance=1cm},
  level 2/.style={sibling distance=1.5cm, level distance=1cm}
  ]
  \node {AND}
    child { node {OR}
      child { node {dog}}
      child { node {AND}
        child { node {cat}}
        child { node {mouse}}
      }
    }
    child { node {NOT}
      child { node {giraffe}}
    };
\end{tikzpicture}
\end{center}

And the computational graph for combining the scores is,
\begin{center}
\begin{tikzpicture}[
  edge from parent/.style={draw, thick, edge from parent fork down},
  level 1/.style={sibling distance=2cm, level distance=1cm},
  level 2/.style={sibling distance=1.5cm, level distance=1cm}
  ]
  \node {$\OPAND$}
    child { node {$\OPOR$}
      child { node {$s_\text{dog}$}}
      child { node {$\OPAND$}
        child { node {$s_\text{cat}$}}
        child { node {$s_\text{mouse}$}}
      }
    }
    child { node {$\OPNOT$}
      child { node {$s_\text{giraffe}$}}
    };
\end{tikzpicture}
\end{center}

With our actual choice of operators this looks like,
\begin{center}
\begin{tikzpicture}[
  edge from parent/.style={draw, thick, edge from parent fork down},
  level 1/.style={sibling distance=2cm, level distance=1cm},
  level 2/.style={sibling distance=1.5cm, level distance=1cm}
  ]
  \node {$*$}
    child { node {$+$}
      child { node {$s_\text{dog}$}}
      child { node {$*$}
        child { node {$s_\text{cat}$}}
        child { node {$s_\text{mouse}$}}
      }
    }
    child { node {$1-\cdot$}
      child { node {$s_\text{giraffe}$}}
    };
\end{tikzpicture}
\end{center}
Written out as a formula this gives the final retrieval score as
$$s = (s_\text{dog} + s_\text{cat}*s_\text{mouse})*(1-s_\text{giraffe}).$$
\section{Additional Results}

\begin{table}[t]
\centering
\begin{tabular}{ccc}
\toprule
 Embedding Model & \textbf{Base} & \textbf{\Method}\\
\midrule
\texttt{NV-Embed-V1} & 0.74 & \textbf{0.97 }\\
\texttt{text-embedding-v3-large} & 0.71 & \textbf{0.97} \\
\texttt{text-embedding-v3-small} & 0.71 & \textbf{0.97} \\
\bottomrule
\end{tabular}
\caption{nDCG@10 results on synthetic data, broken down by embedding model.}
\label{tab:synth_data_embd_model}
\end{table}

\begin{table}
\setlength\tabcolsep{4pt}
\centering
\begin{tabular}{lrrr}
\toprule
& \textbf{NFCorpus} & \textbf{SciFact} & \textbf{ArguAna} \\
\midrule
Accuracy & 87.29 & 86.25 & 85.1 \\
\bottomrule
\end{tabular}
\caption{Accuracy of question transformation into logical queries.}
\label{tab:transf_acc}
\end{table}

\paragraph{Breakdown by Embedding Model} Table~\ref{tab:synth_data_embd_model} breaks down the synthetic data results by embedding model. We see that there is little difference between the embedding models we considered.

\paragraph{Accuracy of Query Transformation} We investigated how accurate Llama3-70b was at transforming natural language questions into logical queries. For all the questions we generated for the three datasets in our experiments--NFCropus, SciFact and ArguaAna--we asked human annotators to assess whether or not the logical query generated by the LLM accurately reflected the original question. As can be seen in Table~\ref{tab:transf_acc}, the generated logical queries tend to capture the original question quite well, with accuracies of 85\% to 87\%. We also examined the queries labelled as inaccurate and found that in the cases we considered the errors tended to be minor. Below we give an example of inaccurate queries from the NFCorpus, SciFact and ArguAna datasets respectively.

\begin{itemize}
    \item \textbf{Example 1} Natural language question:\begin{quote}
        What are some studies on the health benefits of anthocyanins that are not specifically focused on cancer prevention or targeted towards men?
    \end{quote}
    LLM generated query:\begin{quote}
        "Health benefits of anthocyanins" AND (NOT "cancer prevention" OR NOT "targeted towards men")
    \end{quote}
    Here, the original question is ambiguous about which studies should be excluded.
    \item \textbf{Example 2} Natural language question: \begin{quote}
        What are some recent research topics in the fields of neurology and oncology?
    \end{quote} LLM generated query:
    \begin{quote}
        "Recent research in neurology" OR "Current research topics in neurology" AND "Recent advancements in oncology" OR "Current studies in oncology"
    \end{quote}
    Here the \AND should have been an \OR.
    \item \textbf{Example 3} Natural language question: \begin{quote}
        What are some news articles that are not about agricultural development in Africa or Brexit, or are about a topic unrelated to peacekeeping?
    \end{quote}LLM generated query:\begin{quote}
        "News articles" AND NOT ("Agricultural development in Africa" OR "Brexit" OR "Peacekeeping")
    \end{quote}
    Here, "Peacekeeping" should be outside the \NOT clause.
\end{itemize}

\paragraph{Additional Discussion of No-Negative Cases} In Table~\ref{tab:real_data_negatives} we see that most of the gains of \Method~come from cases without negations. In our synthetic experiments (Figure~\ref{fig:and_or}) we show that as the number of terms increases \Method outperforms baselines including on cases without negations. However the improvement is limited when the queries are simpler. We hypothesize that simpler queries are better represented in the training distribution of the retrieval models and are easy enough to process. One thing to note is that because \Method is a modular system, the user can choose to use the base retrieval model if preferred when the extracted logical formulations are simple.

\section{Data Examples}

Below we provide examples of queries from our NFCorpus dataset.

\begin{itemize}
    \item \textbf{Example 1}\begin{quote}
        What are the risk factors for pancreatic cancer, excluding those related to MRSA in swine farms, or what are the benefits of cruciferous compounds in cancer prevention?
    \end{quote}
    \item \textbf{Example 2}\begin{quote}
        What are the health benefits and risks of a vegetarian diet that does not include dairy products, and are there any natural alternatives to dairy that can provide similar nutritional value?
    \end{quote}
    \item \textbf{Example 3}\begin{quote}
        What are some ways to prevent cancer through diet, excluding the effects of xenohormesis mechanisms, and specifically considering the potential benefits of cherries or other foods rich in phenolic compounds?
    \end{quote}
    \item \textbf{Example 4}\begin{quote}
        What are the dietary factors that can help prevent cancer, excluding those related to polycyclic aromatic hydrocarbons?
    \end{quote}
    \item \textbf{Example 5}\begin{quote}
        What are the effects of bioactive compounds on colon or prostate cancer, excluding studies on their mechanisms of action?
    \end{quote}
\end{itemize}

Below, we give the corresponding logical queries generated by the LLM,

\begin{itemize}
    \item \textbf{Example 1}\begin{quote}
        ("Risk factors for pancreatic cancer" AND NOT "MRSA in swine farms") OR "Benefits of cruciferous compounds in cancer prevention"
    \end{quote}
    \item \textbf{Example 2}\begin{quote}
        "Health benefits of a vegetarian diet without dairy products" AND "Risks of a vegetarian diet without dairy products" AND ("Natural alternatives to dairy products" OR "Plant-based alternatives to dairy products") AND ("Nutritional value of dairy products" OR "Nutritional benefits of dairy alternatives")
    \end{quote}
    \item \textbf{Example 3}\begin{quote}
        "Dietary prevention of cancer" AND NOT "xenohormesis" AND ("cherries" OR "foods rich in phenolic compounds")
    \end{quote}
    \item \textbf{Example 4}\begin{quote}
        "Dietary factors that help prevent cancer" AND NOT "Polycyclic aromatic hydrocarbons"
    \end{quote}
    \item \textbf{Example 5}\begin{quote}
        "Effects of bioactive compounds on colon cancer" OR "Effects of bioactive compounds on prostate cancer" AND NOT "Mechanisms of action of bioactive compounds"
    \end{quote}
\end{itemize}

\section{Reformulation Prompts}
Here we give the prompts used for query reformulation in our reasoning methods from Sec.~\ref{sec:real_data}.

\paragraph{BRIGHT Reasoning prompt}

\begin{verbatim}
Here is a user query:

\{question\}

(1) Identify the essential question in the query. 
(2) Think step by step to reason about what should be included in the relevant documents. 
(3) Draft an answer.
\end{verbatim}

\paragraph{Logical Formula Prompt}
\begin{verbatim}
I have a document retrieval system that processes logical queries. 
These queries can be of the form,
"term1" AND "term2" OR "term3" AND NOT "term4"

The meaning of the operators AND, OR and NOT should be obvious:
- AND means the retrieved document should be related to both terms
- OR means the retrieved document can be related to either term
- NOT means the retrieved document should not be related to the given term

Given a natural language question from a user, I want to use the retrieval system to 
gather documents that contain information relevant to the user's question.
I need you to create suitable logical query to the retrieval system. 
Remember that each of the individual terms can be a keyword, a phrase, a sentence, or even 
a whole document. So don't limit yourselves to keywords.
For example, the following question,
"What is the impact of eating fresh oranges on pancreatic cancer risk, 
and its relationship to stage II diabetes"
Could be answered with the query,
"Impact of fresh orange consumption on pancreatic cancer risk" AND 
"What is the relationship of eating fresh oranges to stage II diabetes?"
A query that only used keywords, such as
"oranges" AND "consumption" AND "pancreatic cancer" AND "stage II diabetes"
would lose much of the meaning of the original question! It's not clear if consumption relates 
to oranges, so a document that talks about consuming figs, and peeling oranges 
would match this query!

Here is the user's question,
\{question\}

Can you come up with a suitable logical query to the retrieval system? Only include the query 
in your answer.
\end{verbatim}

\section{Calibration}

We also experimented with calibrating the scores from the retrieval models before combining them in \Method. On the synthetic data, for a given embedding model and for each term, we generated 20 positive and 20 negative documents using Llama3-70b. This gave us a dataset of documents $x_1,\dots,x_N$, with embedded cosine similarities $s_1,\dots,s_N$ and labels $y_1,\dots,y_N\in\{0,1\}$. We then fit a simple calibration model,
\begin{equation*}
    \hat{y}_i = \sigma((s_i - \tau) * \lambda),
\end{equation*}
using gradient descent. This calibration offered some improvements on our synthetic dataset, as can be seen in Table~\ref{tab:synth_data_calibration}, which is an expanded version of Table~\ref{tab:synth_data} with calibration.

\begin{table}
\centering
\begin{tabular}{lrrrr}
\toprule
& \multicolumn{4}{c}{Number of negations} \\
 & 0 & 1 & 2 & 3 \\
Base & 0.95 & 0.77 & 0.65 & 0.52 \\
\Method & 0.99 & 0.97 & 0.96 & \textbf{1.00} \\
Calibrated \Method & \textbf{1.00} & \textbf{0.98} & \textbf{0.97} & \textbf{1.00} \\
\bottomrule
\end{tabular}
\caption{Expanded version of Table~\ref{tab:synth_data} including calibration.}
\label{tab:synth_data_calibration}
\end{table}

\begin{figure}[!ht]
    \centering
    \begin{minipage}{0.475\textwidth}
        \centering
        \includegraphics[width=0.8\linewidth]{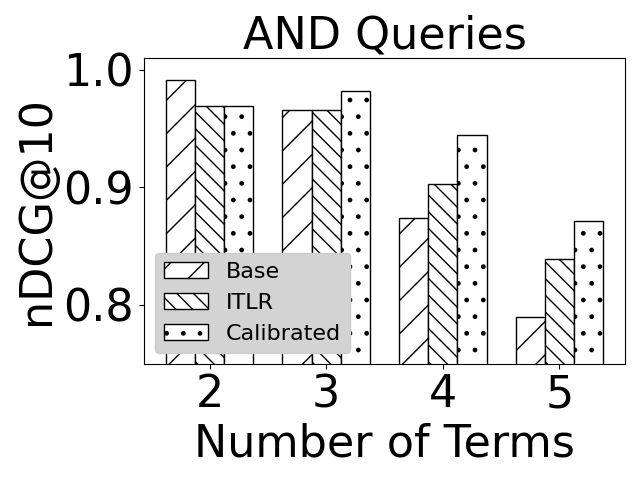}  
        \label{fig:plot1}
    \end{minipage}%
    \begin{minipage}{0.475\textwidth}
        \centering
        \includegraphics[width=0.8\linewidth]{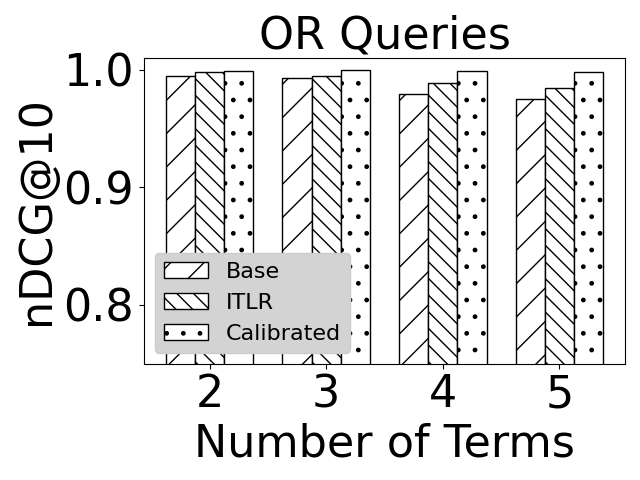}  
        \label{fig:plot2}
    \end{minipage}
    \caption{Expanded version of Fig.~\ref{fig:and_or} including calibration.}
    \label{fig:and_or_calibration}
\end{figure}

\section{Additional Data Details}

Table~\ref{tab:dataset_stats} gives the statistics of the datasets used in our experiments. The datasets were accessed through MTEB \citep{muennighoff2022mteb} under an Apache 2.0 license. These datasets all contain English text.

\begin{table}
\centering
\begin{tabular}{lrr}
\toprule
\textbf{Base Dataset} & \textbf{Corpus Size} & \textbf{Number of Generated Queries} \\
\midrule
NFCorpus & 3,633 & 960 \\
ArguAna & 8,674 & 960 \\
SciFact & 5,183 & 960 \\
\bottomrule
\end{tabular}
\caption{Dataset statistics}
\label{tab:dataset_stats}
\end{table}

\section{Model Details}
Table~\ref{tab:model_params} gives details on the embedding models and LLMs used in our experiments, including parameter counts and how they were accessed.

\begin{table}
\centering
\begin{tabular}{lrrl}
\toprule
\textbf{Model} & \textbf{Number of Model Parameters} & \textbf{Access} & \textbf{License} \\
\midrule
\texttt{Llama3-70b} & $70*10^9$ & API & cc-by-nc-4.0 \\
\texttt{NV-Embed-V1} & $7.85 * 10^9$ & API & llama3 \\
\texttt{text-embedding-3-large} & Undisclosed & API & commercial\\
\texttt{text-embedding-3-small} & Undisclosed & API & commercial \\
\bottomrule
\end{tabular}
\caption{Dataset statistics}
\label{tab:model_params}
\end{table}

\section{Synthetic Data Generation}
\label{sec:app_synth_data}

Consider a query,
\begin{equation*}
    \text{``mouse"} \OR \text{``dog"} \AND \NOT \text{``cat"}.
\end{equation*}

Any document that matches this query can be categorized by the individual terms of the query that it does or does not match. For example, for this query, there are two categories,
\begin{enumerate}
    \item Matches mouse but not cat
    \item Matches dog but not cat
\end{enumerate}
Any of these categories is a list of terms that the document matches, and a list of terms that it doesn't match. This gives an easy way to generate matching documents using Llama3-70b by providing terms that should or shouldn't be matched.

There are many documents that don't match the query, but we want to evaluate against challenging negatives. We can create these by taking one of the categories above and swapping a condition. For example, a positive document will match ``mouse'' but not ``cat''. A hard negative could match ``mouse'' \textit{and} ``cat''. In general, given a list of terms that the document should or shouldn't match, we simply move one term to the opposite category. So a hard negative will match all the terms it should to match the query, except one. Or it will avoid all the terms it should, except one.

Thus, to generate documents for our synthetic data, we first enumerate all the combinations that positive documents should or shouldn't match. We then create all the hard negatives by swapping one of the terms. For each set of terms to match or not match, we create one document. For three term queries, this will result in at most three positive documents. We also keep up two three negative documents. Hence for each query, we generate up to six documents.

\section{Synthetic Queries}
\label{app:synth_queries}

We found that the queries in original datasets used in our experiments are overly simple. For example, in the NFCorpus dataset the queries are the titles of web articles such as "Philippines". To better evaluate the retrieval performance under complex user questions, we created a set of queries based on topics found in the corpus and then used an LLM to evaluate how relevant each passage was to each query. The passages are completely unchanged. This gives us natural and challenging queries with a better labelling of relevant passages, while retaining the complexity of a real world document corpus.

To generate the queries, we extract topics from random documents in the corpus. We then create logical queries using the extracted topics. Finally, we transform these into natural language questions. The Llama3-70b prompts we used are given below.

\paragraph{Topic Extraction Prompt}
\begin{verbatim}
You will be given a document. You need to extract all the salient topics from it. 
The topics should range from general to specific. Here is the document:
{format_doc(doc)}

Please give the salient topics as a list with one topic per line.
Don't include anything else in your answer. 
Sort the topics from most general to most specific.
\end{verbatim}

\paragraph{Query to Question Prompt}
\begin{verbatim}
I have a document retrieval system that processes logical queries. 
These queries can be of the form,
"term1" AND "term2" OR "term3" AND NOT "term4"

The meaning of the operators AND, OR and NOT should be obvious:
- AND means the retrieved document should be related to both terms
- OR means the retrieved document can be related to either term
- NOT means the retrieved document should not be related to the given term

I want to evaluate the performance of a human user to use this retrieval system. 
Given a natural language question, the user needs to come up with a logical query that will best 
retrieve relevant documents.
In order to make a dataset for evaluation, I want to operate in reverse. I have collected 
many logical queries, and I would like to come up with a corresponding natural language question.
Then I can give the question to a user and see how well the recover they original query.

So, given the following logical query, can you come up with such a natural language question? 
Here's the query,
{query}

What question would you come up with? Only include the question in your answer.
\end{verbatim}

\section{Human Annotations}
Human annotators tasked with evaluating the LLM generated queries were paid a fair wage of $\$25$ an hour. They were given the following instructions
\begin{lstlisting}
Instruction for Human Annotator: Logical Expression Validation
Task Overview
You will be given a natural language question and a corresponding logical expression generated by an LLM (Large Language Model). Your task is to determine whether the logical expression accurately represents the intended meaning of the question.
A correct logical expression should:
- Capture the key intent of the question.
- Properly reflect any exclusions, inclusions, or constraints mentioned.
- Maintain the logical relationships between elements.
Evaluation Criteria
1. Accuracy - Does the logical expression correctly interpret the intent of the question?
2. Completeness - Are all relevant aspects of the question included in the logical expression?
3. Exclusions - If the question explicitly excludes something, does the logical expression handle this correctly?
4. Logical Structure - Are the AND, OR, and NOT operators used correctly to reflect the relationships in the question?
If the logical expression is correct, mark it as valid. If incorrect, mark it as invalid and provide an explanation of the error.
Examples
Positive Examples (Correct Expressions)
Example 1:
- Natural Language Question: ``How does vitamin D benefit your health? I already know about bone health, so I want to know other benefits.''
- Parsed Logical Expression: health benefits of vitamin D AND NOT bone health
- Explanation: The logical expression correctly retrieves information about vitamin D's health benefits while excluding bone health, as specified in the question.
Example 2:
- Natural Language Question: ``What are some movies directed by Christopher Nolan, excluding superhero films?"
- Parsed Logical Expression: movies directed by Christopher Nolan AND NOT superhero films
- Explanation: The logical expression correctly filters out superhero films while still retrieving Nolan's other movies.
Example 3:
- Natural Language Question: ``Which laptops have at least 16GB RAM and either an Intel i7 or AMD Ryzen 7 processor?"
- Parsed Logical Expression: laptops AND 16GB RAM AND (Intel i7 OR AMD Ryzen 7)
- Explanation: The expression correctly captures the requirement of 16GB RAM and allows either processor type, as intended.
Negative Examples (Incorrect Expressions)
Example 4:
- Natural Language Question: ``How does vitamin D benefit your health? I already know about bone health, so I want to know other benefits."
- Parsed Logical Expression: health benefits of vitamin D OR NOT bone health
- Error: The use of OR NOT instead of AND NOT changes the meaning. The expression may return results that are completely unrelated to vitamin D.
Example 5:
- Natural Language Question: ``What are some movies directed by Christopher Nolan, excluding superhero films?"
- Parsed Logical Expression: movies directed by Christopher Nolan OR NOT superhero films
- Error: The OR NOT operator incorrectly allows movies that aren't superhero films but might not be directed by Nolan, which is not what the question asks.
Example 6:
- Natural Language Question: ``Which laptops have at least 16GB RAM and either an Intel i7 or AMD Ryzen 7 processor?"
- Parsed Logical Expression: laptops AND (16GB RAM OR Intel i7 OR AMD Ryzen 7)
- Error: The use of OR within the parentheses makes it possible for laptops with only Intel i7 or AMD Ryzen 7 (but less than 16GB RAM) to be included, which is incorrect.
Final Notes
- Pay close attention to negations (NOT). Misplacing them can completely alter the meaning.
- Ensure correct grouping with parentheses. Ambiguities in logic can lead to unintended results.
- Rephrase the natural language question in a structured way before checking the logical expression.
Your accuracy in annotation ensures that the model correctly understands and processes logical constraints in natural language.
\end{lstlisting}

\end{document}